\documentclass[runningheads]{llncs}
\usepackage{graphicx}
\usepackage{float}
\usepackage{multirow}
\usepackage{graphicx}
\usepackage{longtable}
\usepackage{tabularx}
\usepackage{enumitem}
\usepackage{booktabs}
\usepackage{url}
\usepackage{subcaption}
\usepackage{color}
\usepackage{caption}
\usepackage{amsmath,amsfonts,amssymb,epsfig,url,epstopdf,array}
\usepackage{makecell}
\DeclareMathOperator*{\argmax}{arg\,max}

\usepackage{appendix}

\begin{document}
\title{Multi-objective hyperparameter optimization with performance uncertainty\thanks{\mailname{ Alejandro Morales-Hernández}}}
\author{Alejandro Morales-Hernández\inst{1,2,3}\orcidID{0000-0003-0053-4902} \and
Inneke Van Nieuwenhuyse\inst{1,2,3}\orcidID{0000-0003-2759-3726} \and
Gonzalo N\'apoles\inst{4}\orcidID{0000-0003-1936-3701}}
\authorrunning{A. Morales-Hernández et al.}
\institute{Core Lab VCCM, Flanders Make, Limburg, Belgium
\and Research Group Logistics, Hasselt University, Agoralaan Gebouw D, Diepenbeek, 3590, Limburg, Belgium 
\and Data Science Institute, Hasselt University, Agoralaan Gebouw D , Diepenbeek, 3590, Limburg, Belgium \\\email{\{alejandro.moraleshernandez, inneke.vannieuwenhuyse\}@uhasselt.be}
\and Department of Cognitive Science \& Artificial Intelligence, Tilburg University, The Netherlands \\\email{g.r.napoles@uvt.nl}}
\maketitle              % typeset the header of the contribution
\begin{abstract}
The performance of any Machine Learning algorithm is impacted by the choice of its hyperparameters. As training and evaluating a ML algorithm is usually expensive, the hyperparameter optimization (HPO) method needs to be computationally efficient to be useful in practice. Most of the existing approaches on multi-objective HPO use evolutionary strategies and metamodel-based optimization. However, few methods account for uncertainty in the performance measurements. This paper presents results on multi-objective HPO with uncertainty on the performance evaluations of the ML algorithms. We combine the sampling strategy of Tree-structured Parzen Estimators (TPE) with the metamodel obtained after training a Gaussian Process Regression (GPR) with heterogeneous noise. Experimental results on three analytical test functions and three ML problems show the improvement in the hypervolume obtained, when compared with HPO using stand-alone multi-objective TPE and GPR. 

\keywords{hyperparameter optimization \and multi-objective optimization \and Bayesian optimization \and uncertainty}
\end{abstract}
\section{Introduction}
\label{sec:introduction}

In Machine Learning (ML), an hyperparameter is a parameter that needs to be specified before training the algorithm: it influences the learning process, but it is not optimized as part of the training algorithm. The time needed to train a ML algorithm with a given hyperparameter configuration on a given dataset may already be substantial, particularly for moderate to large datasets, so the HPO algorithm should be as efficient as possible in detecting the optimal hyperparameter setting.

Many of the current algorithms in the literature focus on optimizing a single (often error-based) objective \cite{bergstra2011algorithms,snoek2012practical,li2017hyperband}. In practical applications, however, it is often required to consider the trade-off between two or more objectives, such as the error-based performance of a model and its resource consumption \cite{igel2005multi}, or objectives relating to different types of error-based performance measures \cite{horn2016multi}. The goal in multi-objective HPO is to obtain the \emph{Pareto-optimal} solutions, i.e., those hyperparameter values for which none of the performance measures can be improved without negatively affecting any other. 

In the literature, most HPO approaches take a deterministic perspective using the mean value of the performance observed in subsets of data (cross validation protocol). However, depending on the chosen sets, the outcome may differ: a single HP configuration may thus yield different results for each performance objective, implying that the objective contains uncertainty (hereafter referred to as \emph{noise}). We conjecture that a HPO approach that considers this uncertainty will outperform alternative approaches that assume the relationships to be deterministic. Stochastic algorithms (such as \cite{binois2019replication,gonzalez2020multiobjective}) can potentially be useful for problems with heterogeneous noise (the noise level varies from one setting to another). To the best of our knowledge, such approaches have not yet been studied in the context of HPO optimization. The main contributions of our approach include:

\begin{itemize}
\item Multi-objective optimization using a Gaussian Process Regression (GPR) surrogate that explicitly accounts for the heterogeneous noise observed in the performance of the ML algorithm.

\item The selection of infill points according to the sampling strategy of multi-objective TPE (MOTPE), and the maximization of an infill criterion. This method allows sequential selection of hyperparameter configurations that are likely to be non-dominated, and that yield the largest expected improvement in the Pareto front.
\end{itemize}

The remainder of this article is organized as follows. Section \ref{sec:metamodel_moo} discusses the basics of GPR and MOTPE. Section \ref{sec:algorithm} presents the algorithm. Section \ref{sec:simulations} describes the experimental setting designed to evaluate the proposed algorithm, and Section \ref{sec:results} shows the results. Finally, Section \ref{sec:conclusions} summarizes the findings and highlights some future research directions.

\section{GPR and TPE: Basics}
\label{sec:metamodel_moo}

Gaussian Process Regression (GPR) (also referred to as \textit{kriging}, \cite{williams2006gaussian}) is commonly used to model an unknown target function. The function value prediction at an unsampled point $\mathbf{x}^{(*)}$ is obtained through the conditional probability $P(f(\mathbf{x}^{(*)})|\textbf{X,Y})$ that represents how likely the response $f(\mathbf{x}^{(*)})$ is, given that we observed the target function at $n$ input locations $\mathbf{x}^{(i)}, i=1, \dots, n$ (contained in matrix $\textbf{X}$), yielding function values $\mathbf{y}^{(i)}, i=1, \dots, n$ (contained in matrix $\textbf{Y}$) that may or may not be affected by noise. Ankenman et al. \cite{Anke10} provides a GPR model (referred to as \emph{stochastic kriging}) that takes into account the heterogeneous noise observed in the data, and models the observed response value in the \textit{r}-th replication at design point $\mathbf{x}^{(i)}$ as:

\begin{equation}
\label{eq:stochastic_regression}
    f_r(\mathbf{x}^{(i)})=m(\mathbf{x}^{(i)}) + M(\mathbf{x}^{(i)})+\epsilon_r(\mathbf{x}^{(i)})
\end{equation}

\noindent where $m(\mathbf{x})$ represents the mean of the process, $M(\mathbf{x})$ is a realization of a Gaussian random field with mean zero (also referred to as the \emph{extrinsic uncertainty} \cite{Anke10}), and $\epsilon_r(\mathbf{x}^{(i)})$ is the \textit{intrinsic uncertainty} observed in replication $r$. Popular choices for $m(\mathbf{x})$ are $m(\mathbf{x})=\sum_h\beta_h f_h(\mathbf{x})$ (where the $f_h(\mathbf{x})$ are known linear or nonlinear functions of $\mathbf{x}$, and the $\beta_h$ are unknown coefficients to be estimated), $m(\mathbf{x})=\beta_0$ (an unknown constant to be estimated), or $m(\mathbf{x})=0$. $M(\mathbf{x})$ can be seen as a function, randomly sampled from a space of functions that, by assumption, exhibit spatial correlation according to a covariance function (also referred to as \emph{kernel}). 

Whereas GPR models the probability distribution of $f(\mathbf{x})$ given a set of observed points ($P(f(\mathbf{x})|\mathbf{X,Y})$), TPE tries to model the probability of sampling a point that is directly associated to the set of observed responses ($P(\mathbf{x}|\mathbf{X,Y})$) \cite{bergstra2011algorithms}. TPE defines $P(\mathbf{x}|\mathbf{X,Y})$ using two densities:

\begin{equation}
\label{eq:tpe}
    P(\mathbf{x}|\mathbf{X,Y}) =\left\{\begin{matrix}
 l(x)& if \quad f(x) < y^*, \mathbf{x} \in \mathbf{X} \\ 
 g(x)& \quad o.w
\end{matrix}\right.
\end{equation}

\noindent where $l(x)$ is the density estimated  using the points ${\mathbf{x}^{(i)}}$ for which $f(\mathbf{x}^{(i)})< y^*$, and $g(x)$ is the density estimated using the remaining points. The value $y^*$ is a user-defined quantile $\gamma$ (splitting parameter of Algorithm 1 in \cite{ozaki2020multiobjective}) of the observed $f(\mathbf{x})$ values, so that $P(f(\mathbf{x}) < y^*)=\gamma$. Here, we can see $l$ as the density of the hyperparameter configurations that may have the best response. A multi-objective implementation of TPE (MOTPE) was proposed by \cite{ozaki2020multiobjective}; this multi-objective version splits the known observations according to their nondomination rank. Contrary to GPR, neither TPE nor MOTPE provide an estimator of the response at unobserved hyperparameter configurations.

\section{Proposed algorithm}
\label{sec:algorithm}

\begin{figure}[!htbp]
\centering
\includegraphics[width=12.5cm]{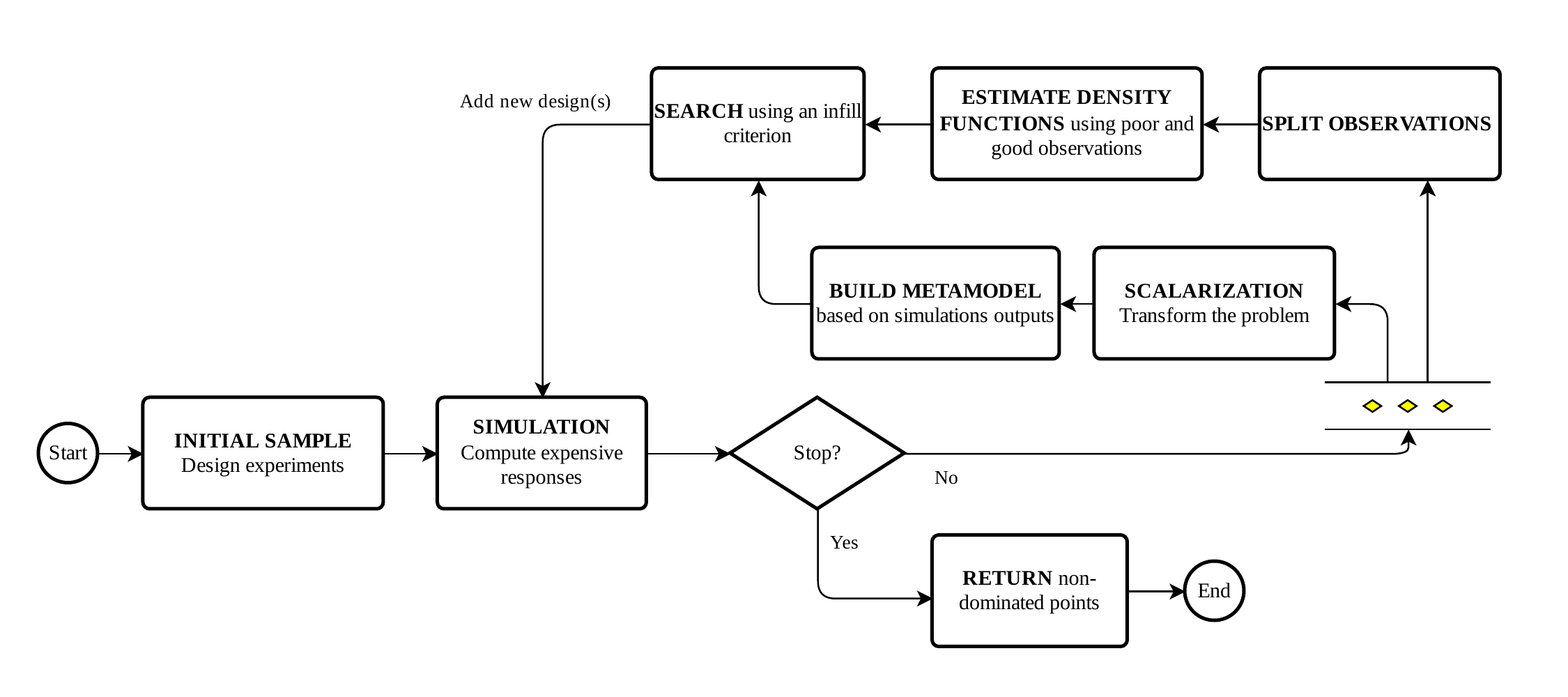} 
\caption{Proposed multi-objective HPO using GPR with heterogeneous noise and TPE to sample the search space}
\label{fig:algorithm}
\end{figure}

The algorithm (Figure \ref{fig:algorithm}) starts by evaluating an initial set of hyperparameter vectors through a Latin hypercube sample; simulation replications are used to estimate the objective values at these points. We then perform two processes in parallel. On the one hand, we use the augmented Tchebycheff scalarization function \cite{knowles2006parego} (with a random combination of weights) to transform the multiple objectives into a single objective using these training data. Throughout this article, we will assume that the individual objectives need to be minimized; hence, the resulting scalarized  objective function also needs to be minimized. We then train a (single) stochastic GPR metamodel on these scalarized objective function outcomes; the replication outcomes are used to compute the variance of this scalarized objective. 

At the same time, we perform the splitting process used by \cite{ozaki2020multiobjective} to divide the hyperparameter vectors into two subsets (those yielding ``good'' and ``poor'' observations) to estimate the densities $l(x)$ and $g(x)$ for each separate input dimension (Eq. \ref{eq:tpe}). To that end, our approach uses a greedy selection according to the nondomination rank of the observations, and controlled by the parameter $\gamma$ \footnote{Notice that both in \cite{ozaki2020multiobjective} and in our algorithm, the parameter $\gamma$ represents a percentage of the known observations that may be considered as ``good''.}. The strategy thus preferably selects the HP configurations with highest nondomination rank to enter in the "good" subset.

Using the densities $l(x)$, we randomly select a candidate set of $n_c$ configurations for each input dimension. These individual values are sorted according to their log-likelihood ratio $log \frac{l(x)}{g(x)}$, such that the higher this score, the larger the probability that the input value is sampled under $l(\mathbf{x}_i)$ (and/or the lower the probability under $g(\mathbf{x}_i)$). Instead of selecting the single configuration with highest score on each dimension (as in \cite{bergstra2011algorithms,ozaki2020multiobjective}), we compute the aggregated score $AS(\mathbf{x})= \sum_{i=1}^{d} log \frac{l(x_i)}{g(x_i)}$ for each configuration, and select the one that maximizes the \emph{Modified Expected Improvement} (MEI) \cite{quan2013simulation} of the scalarized objective function in the set of configurations $Q$ with an aggregated score greater than zero (see Eq. \ref{eq:aggregated_selection}).

\vspace{-5mm}
\begin{equation}
    \label{eq:aggregated_selection}
        \argmax_{\mathbf{q} \in Q} \; (\widehat{Z}_{\min}-\widehat{Z}_{\mathbf{q}})\Phi(\frac{\widehat{Z}_{\min}-\widehat{Z}_\mathbf{q}}{\widehat{s}_\mathbf{q}})+\widehat{s}_\mathbf{q}\phi(\frac{\widehat{Z}_{\min}-\widehat{Z}_\mathbf{q}}{\widehat{s}_\mathbf{q}}) \;, Q = \{ \mathbf{x} \, | \, AS(\mathbf{x}) > 0\}
\end{equation}

where $\widehat{Z}_{\min}$ is the stochastic kriging prediction at $\mathbf{x}_{\min}$ (i.e. the hyperparameter configuration with the lowest sample mean among the already known configurations), $\phi(\cdot)$ and $\Phi(\cdot)$ are the standard normal density and standard normal distribution function respectively, the $\widehat{Z}_\mathbf{q}$ is the stochastic kriging prediction at configuration $\mathbf{q}$, and $\widehat{s}_\mathbf{q}$ is the ordinary kriging standard deviation for that configuration \cite{zhan2020expected}. The search using MEI focuses on new points located in promising regions (i.e., with low predicted responses; recall that we assume that the scalarized objective need to be minimized),  or in regions with high metamodel uncertainty (i.e., where little is known yet about the objective function). Consequently, the sampling behavior automatically  trades off  exploration and exploitation of the configuration search space.

Once a new hyperparameter configuration has been selected as infill point, the ML algorithm is trained on this configuration, yielding (again) noisy estimates of the performance measures. Following this infill strategy, we choose that configuration for which we expect the biggest  improvement in the scalarized objective function, among the configurations that are likely to be non-dominated.

\section{Numerical simulations}
\label{sec:simulations}

In this section, we evaluate the performance of the proposed algorithm for solving multi-objective optimization problems (GP\_MOTPE), comparing the results with those that would be obtained by using GP modelling and MOTPE individually. In a first experiment, we analyze the performance on three well-known bi-objective problems (ZDT1, WFG4 and DTLZ7 with input dimension $d=5$; see \cite{huband2006review}), to which we add artificial heterogeneous noise (as in \cite{gonzalez2020multiobjective}). More specifically, we obtain noisy observations $\widetilde{f}_{p}^j(\mathbf{X}_i)=f_j(\mathbf{X}_i) + \epsilon_p(\mathbf{X}_i), p=\{1, \dots, r\}, j=\{1, \dots, m\}$, with $\epsilon_p(\mathbf{X}_i) \sim \mathcal{N}(0, \tau_j(\mathbf{X}_i))$. The standard deviation of the noise ($\tau_j(\mathbf{X})$) varies for each objective between $0.01 \times \Omega^j$ and $0.5 \times \Omega^j$, where  $\Omega^j$  is the range of objective $j$. In between these limits, $\tau_j(\mathbf{X})$ decreases linearly with the objective value: $\tau_j(\mathbf{X}) = a_j(f_j(\mathbf{X}) + b_j), \forall j \in \{1, \dots, m\}$, where $a$ and $b$ are the linear coefficients obtained from the noise range \cite{jalali2017comparison}.

\vspace{-8mm}
\begin{table}[!hbtp]
\centering
\small
\caption{Details of the ML datasets }
\label{tab:datasets}
\begin{tabular}{lcc}\\
\hline
\multicolumn{1}{c}{\textbf{Dataset}} & \multicolumn{1}{c}{\textbf{ID}} & \multicolumn{1}{c}{\textbf{Inst. (Feat.)}} \\ \hline
Balance-scale & 997 & 625 (4) \\
Optdigits & 980 & 5620 (64)\\
Stock & 841 & 950 (9)\\
Pollen & 871 & 6848 (5)\\
Sylvine & 41146 & 5124 (20) \\
Wind & 847 & 6574 (14) \\
\hline	
\end{tabular}
\quad
\begin{tabular}{lcc}\\
\hline
\multicolumn{1}{c}{\textbf{Dataset}} & \multicolumn{1}{c}{\textbf{ID}} & \multicolumn{1}{c}{\textbf{Inst. (Feat.)}} \\ \hline
Delta\_ailerons & 803 & 7129 (5) \\
Heart-statlog & 53 & 270 (13) \\
Chscase\_vine2 & 814 & 468 (2) \\
Ilpd & 41945 & 583 (10) \\
Bodyfat & 778 & 252 (14) \\
Strikes & 770 & 625 (6) \\ \hline
\end{tabular}
\end{table}
\vspace{-5mm}

In a second experiment, we test the algorithm on a number of OpenML datasets, shown in  Table \ref{tab:datasets}. We optimize five hyperparameters for a simple (one hidden layer) Multi-Layer Perceptron (MLP), two for a support vector machine (SVM), and five for a Decision Tree (DT) (see Appendix A). In each experiment, the goal is to find the HPO configurations that minimize classification error while simultaneously maximizing recall. In all experiments, we used $20\%$ of the initial dataset as test set, and the remainder for HPO. We apply stratified \emph{k-fold cross-validation} ($k=10$) to evaluate each hyperparameter configuration. 

We used a fixed, small number of iterations (100) as a stopping criterion in all algorithms; this keeps optimization time low, and resembles real-world optimization settings where limited resources (e.g., time) may exist. Table \ref{tab:experiments} summarizes the rest of the parameters used in the experiments.

\begin{longtable}[H]{p{2cm}>{\centering}p{2.5cm}>{\centering}p{2cm}>{\centering}p{2cm}>{\centering\arraybackslash}p{1.5cm}} 
\caption{Summary of the parameters for the experiments}
\label{tab:experiments}
\\ \hline
\multicolumn{1}{c}{\textbf{Setting}} & \multicolumn{1}{c}{\textbf{Problem}} & \multicolumn{1}{c}{\textbf{GP}} & \multicolumn{1}{c}{\textbf{MOTPE}} & \multicolumn{1}{c}{\textbf{GP\_MOTPE}} \\ \hline
\endfirsthead
\hline
\multicolumn{1}{c}{\textbf{Setting}} & \multicolumn{1}{c}{\textbf{Problem}} & \multicolumn{1}{c}{\textbf{GP}} & \multicolumn{1}{c}{\textbf{MOTPE}} & \multicolumn{1}{c}{\textbf{GP\_MOTPE}} \\ \hline
\endhead
\hline \multicolumn{4}{r}{{Continued on next page}} \\ 
\endfoot
\endlastfoot

Initial design & Analytical fcts & \multicolumn{3}{c}{LHS: $11d-1$} \\ 
\multicolumn{1}{c}{} & HPO & \multicolumn{3}{c}{Random sampling: $11d-1$} \\ \hline

Replications & Analytical fcts & \multicolumn{3}{c}{50} \\ 
\multicolumn{1}{c}{} & HPO & \multicolumn{3}{c}{10} \\ \hline

\multicolumn{2}{l}{Acquisition function} & $\mathrm{MEI}$ & $\mathrm{EI_{TPE}}$ & $\mathrm{MEI}$ \\ \hline

\multicolumn{2}{l}{Acquisition function optimization} & \textbf{PSO}* & \multicolumn{2}{p{5cm}}{Maximization on a candidate set } \\ \hline

\multicolumn{2}{l}{Number of candidates  to sample} & \multicolumn{1}{c}{-} & \multicolumn{2}{c}{$n_c=1000$ , $\gamma=0.3$} \\ \hline

\multicolumn{2}{l}{Kernel} & Gaussian & \multicolumn{1}{c}{-} & Gaussian \\ \hline
\multicolumn{5}{p{12cm}}{ * PSO algorithm (Pyswarm library): swarm size = 300, max iterations = 1800, cognitive parameter=0.5, social parameter=0.3, and inertia=0.9 } \\
\end{longtable}

\section{Results}
\label{sec:results}

\begin{figure*}[!hbtp]
\centering
\begin{subfigure}[b]{0.3\textwidth}
\includegraphics[width=4cm,height=3cm]{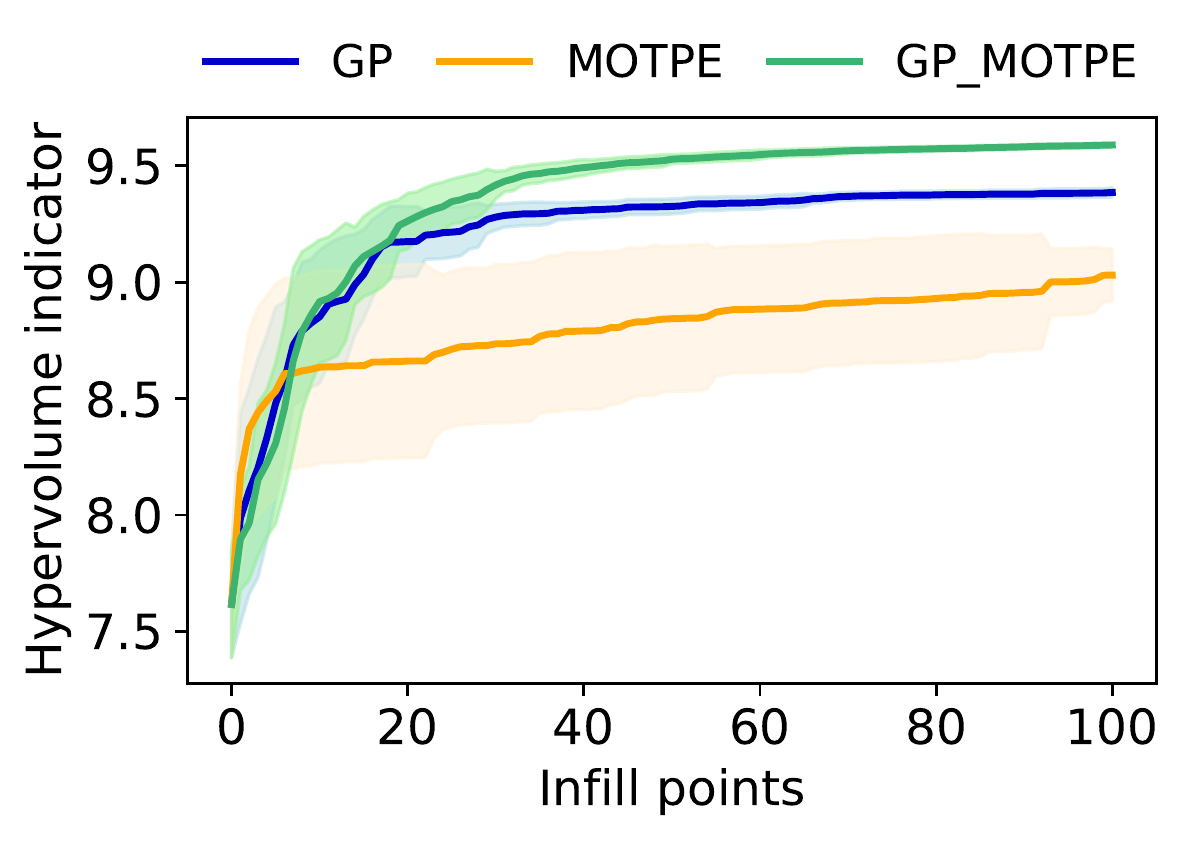}
\caption{ZDT1 $ref=[1, 10]$}
\label{fig:hv_ZDT1}
\end{subfigure}
\hfill
\begin{subfigure}[b]{0.3\textwidth}
\includegraphics[width=4cm]{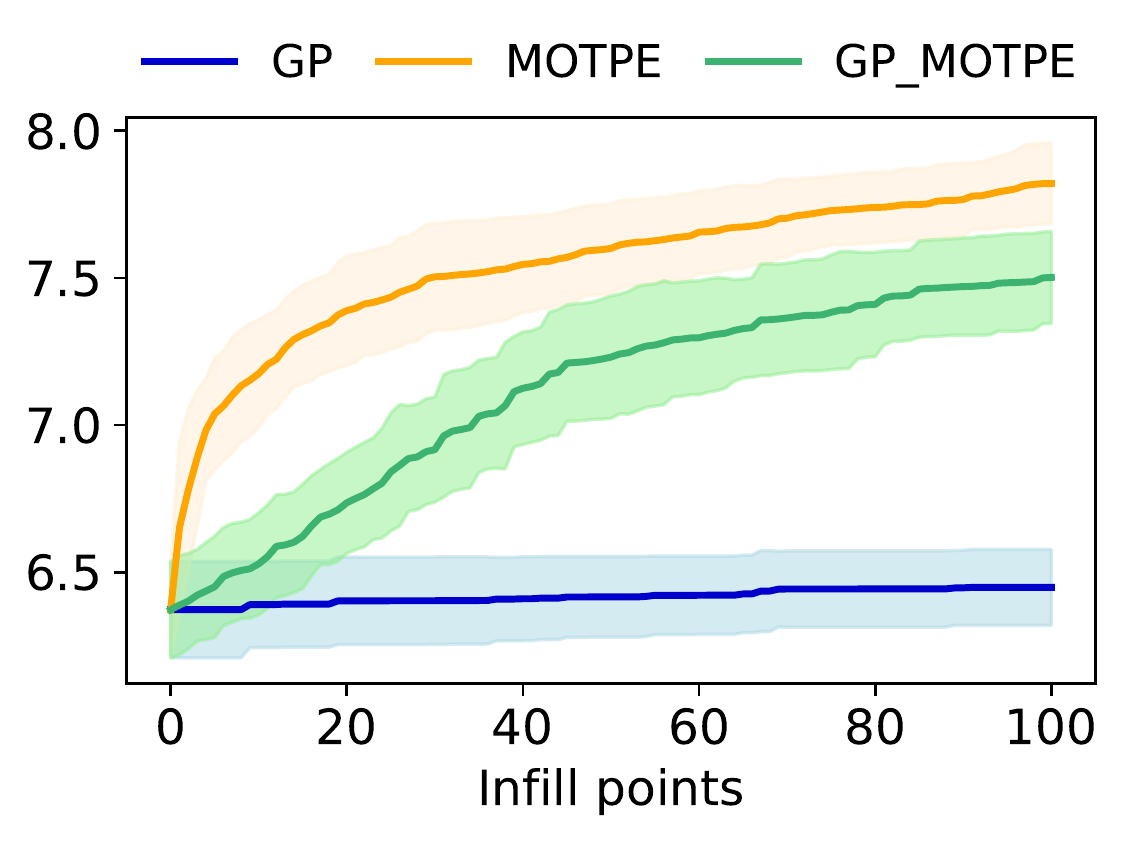}
\caption{WFG4 $ref=[3, 5]$}
\label{fig:hv_WFG4}
\end{subfigure}
\hfill
\begin{subfigure}[b]{0.3\textwidth}
\centering
\includegraphics[width=4cm]{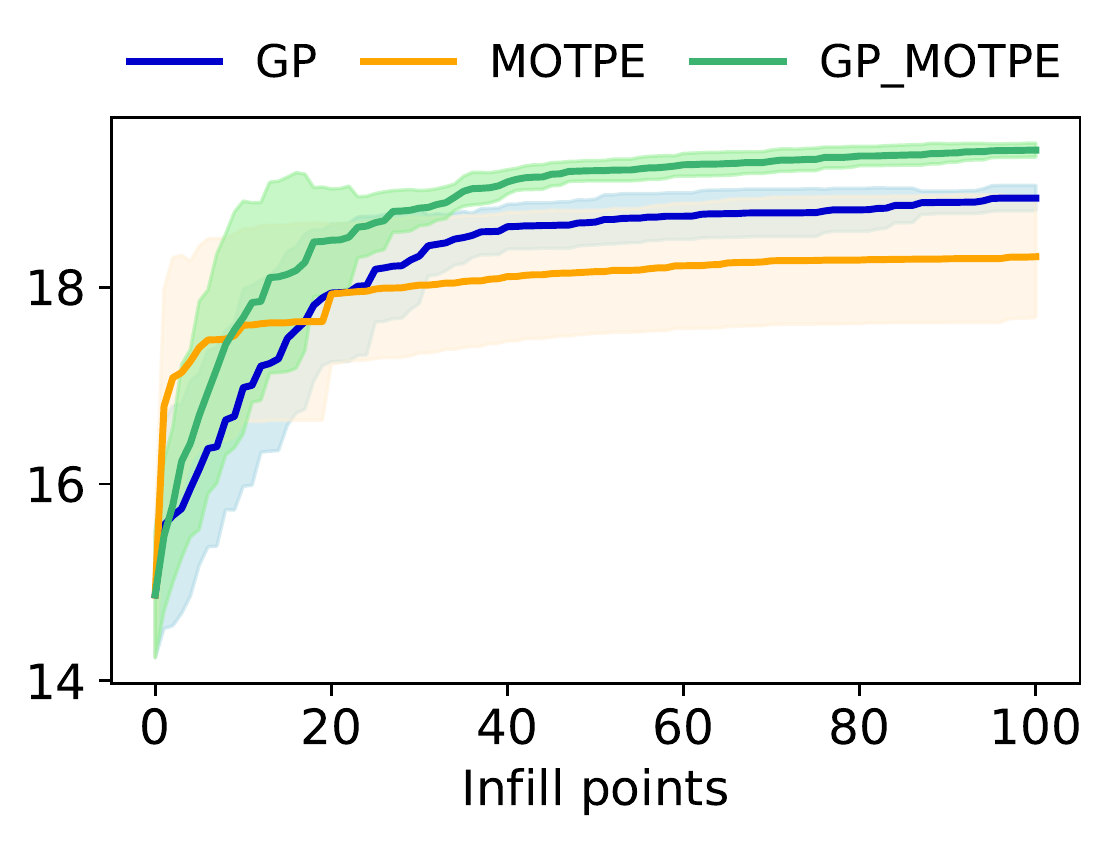}
\caption{DTLZ7 $ref=[1, 23]$}
\label{fig:hv_DTLZ7}
\end{subfigure}

\caption{Hypervolume evolution during the optimization of the analytical test functions. Shaded area represents $mean \pm std$ of 13 macro-replications. Captions contain the reference point used to compute the hypervolume indicator}
\label{fig:hv_test_function}
\end{figure*}

Figure \ref{fig:hv_test_function} shows the evolution of the hypervolume indicator during the optimization of the analytical test functions. The combined algorithm GP\_MOTPE yields a big improvement over both GP and MOTPE algorithms for the ZDT1 and DTLZ7 functions, reaching a superior hypervolume already after a small number of iterations. Results also show that for ZDT1 and DTLZ7, the standard deviation on the final hypervolume obtained by GP and GP\_MOTPE is small, which indicates that a Pareto front of similar quality is obtained regardless of the initial design. MOTPE, by contrast, shows higher uncertainty in the hypervolume results at the end of the optimization. For the concave Pareto front of WFG4, MOTPE provides the best results, while GP\_MOTPE still outperforms GP. 

% \vspace{-5mm}
\begin{figure*}[!hbtp]
\centering
\begin{subfigure}[b]{0.3\textwidth}
\includegraphics[width=3.9cm]{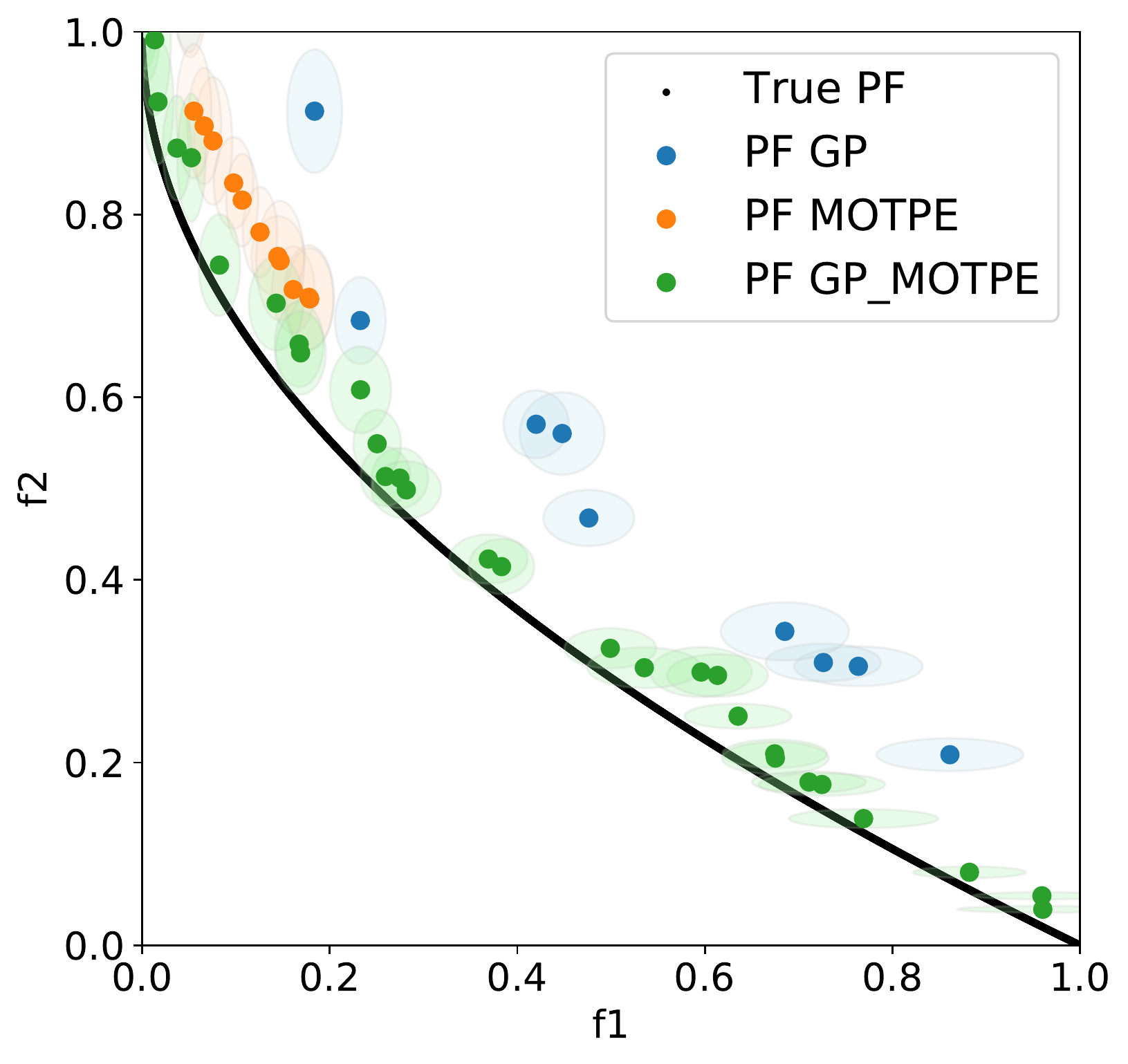}
\caption{ZDT1}
\label{fig:pf_ZDT1}
\end{subfigure}
\hfill
\begin{subfigure}[b]{0.3\textwidth}
\includegraphics[width=3.8cm]{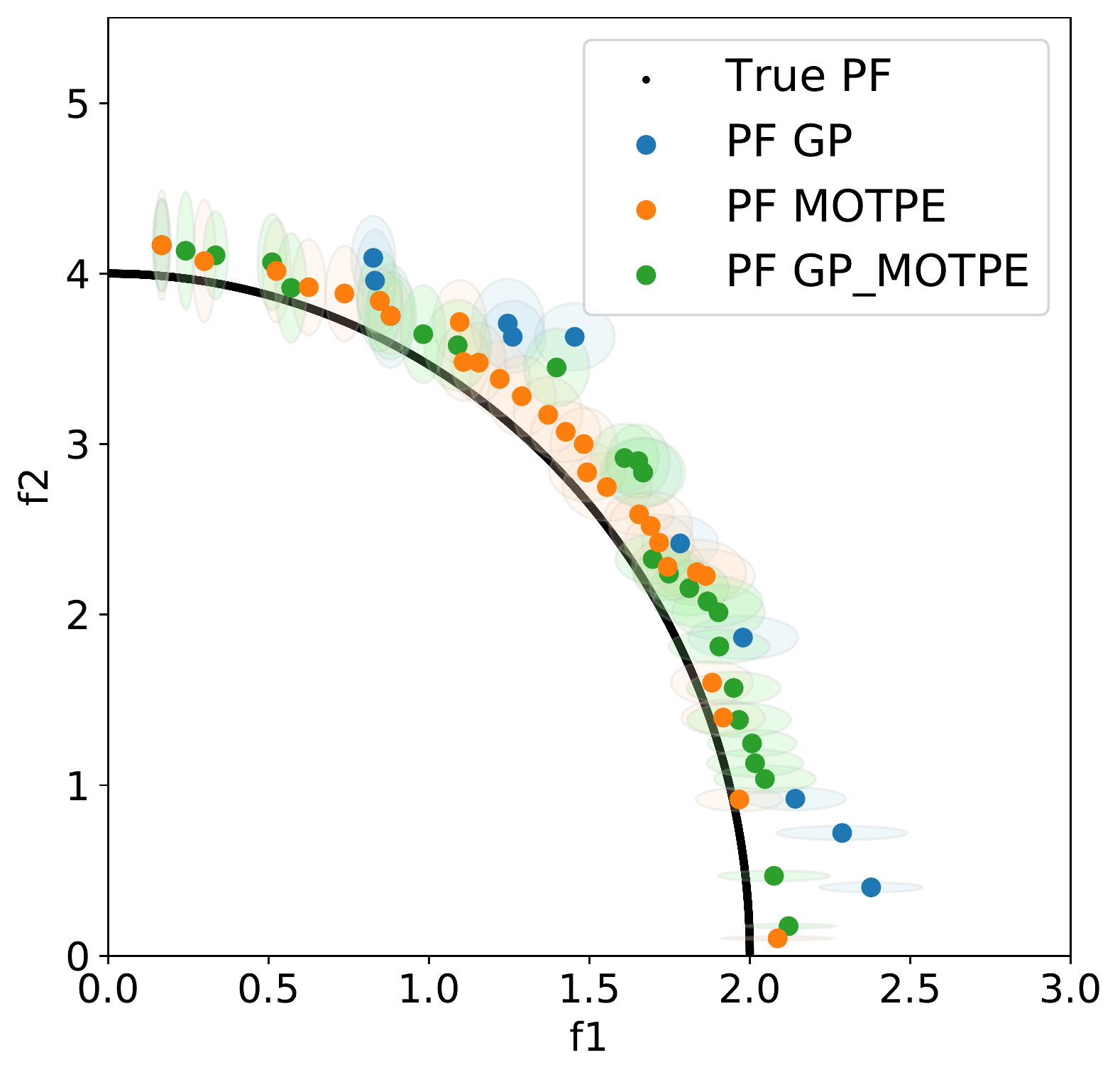}
\caption{WFG4}
\label{fig:pf_WFG4}
\end{subfigure}
\hfill
\begin{subfigure}[b]{0.3\textwidth}
\centering
\includegraphics[width=3.9cm]{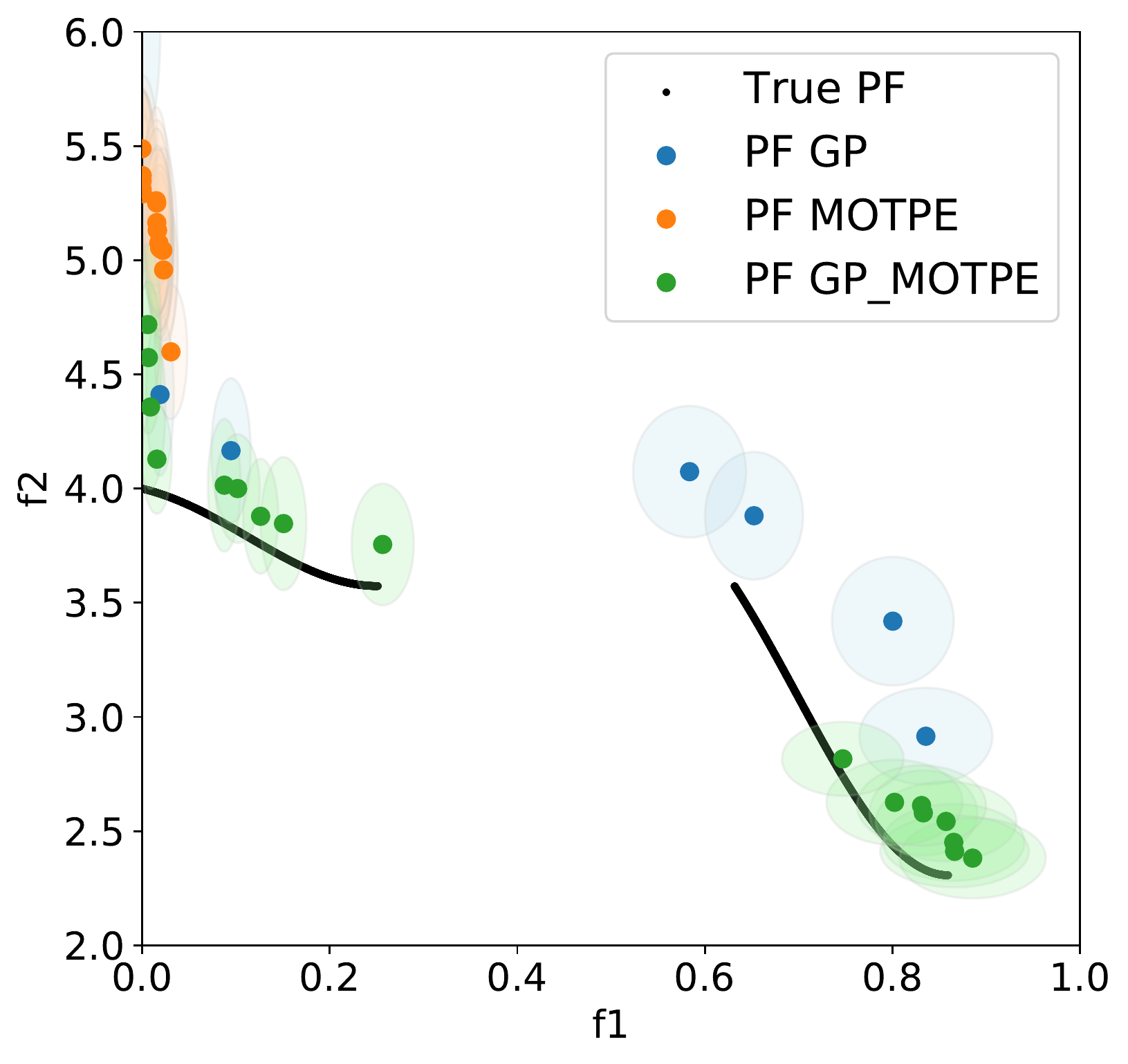}
\caption{DTLZ7}
\label{fig:pf_DTLZ7}
\end{subfigure}

\caption{Observed Pareto front (PF) obtained at the end of a single macroreplication, for the analytical test functions. The uncertainty of each solution is shown by a shaded ellipse, and reflects the $mean \pm std$ of the simulation replications.}
\label{fig:pf_test_function}
\end{figure*}

Table \ref{tab:HV_ML} shows the average rank of the optimization algorithms according to the hypervolume indicator. The experiments did not highlight significant differences between GP\_MOTPE, GP and MOTPE ($p\_value = 0.565 > 0.05$ for the non-parametric Friedman test where $H_0$ states that the mean hypervolume of the solutions is equal). However, GP\_MOTPE has the lowest average rank in the validation set, indicating that on average, the Pareto front obtained with our algorithm tends to outperform those found by GP and MOTPE individually, yielding a larger hypervolume.

% \vspace{-5mm}
% \vspace{-5mm}

\begin{figure*}[!hbtp]
\centering
\begin{subfigure}[b]{0.3\textwidth}
\includegraphics[width=4cm]{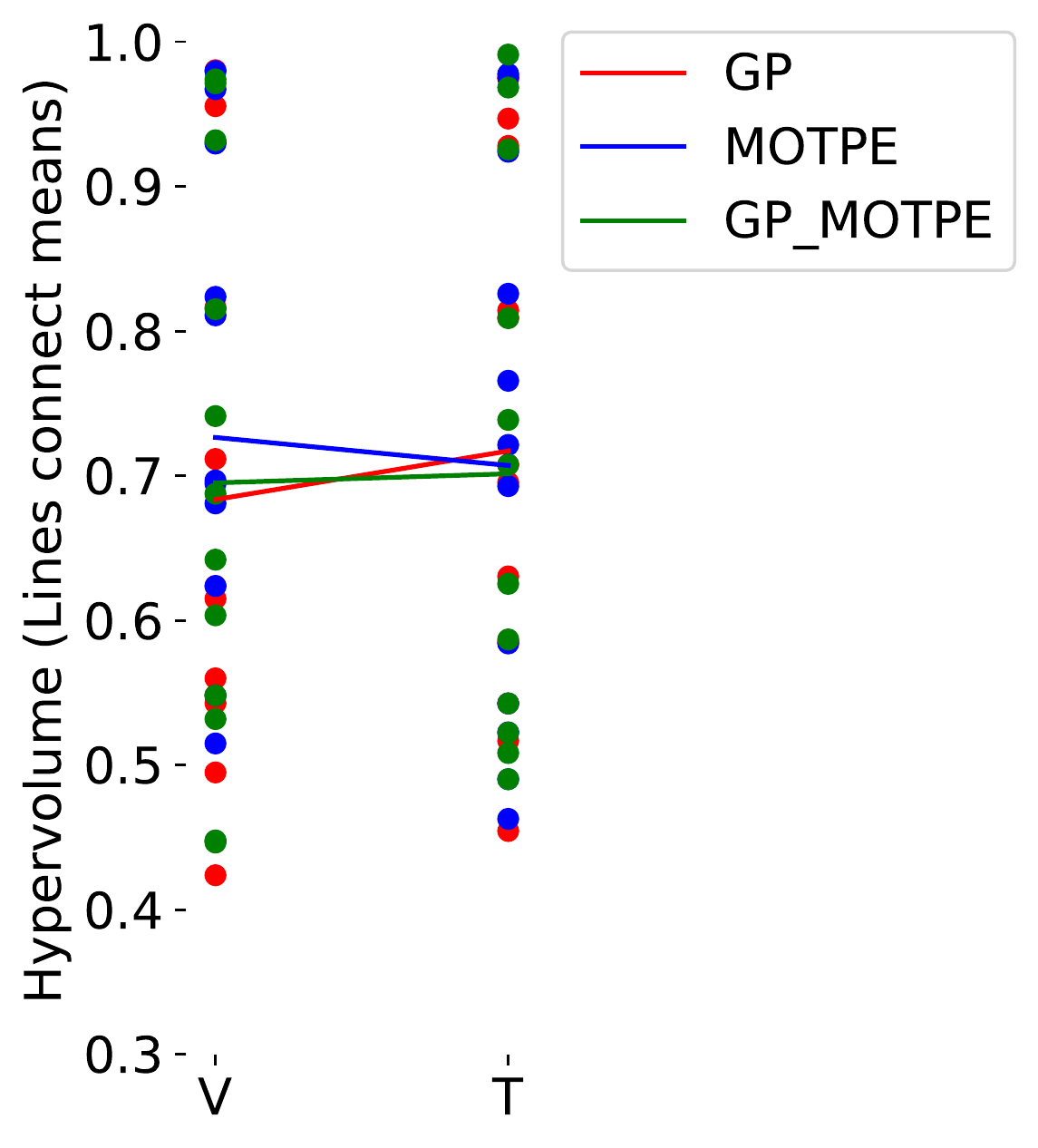}
\caption{MLP}
\label{fig:hv_vt_mlp}
\end{subfigure}
\hfill
\begin{subfigure}[b]{0.3\textwidth}
\includegraphics[width=3.9cm]{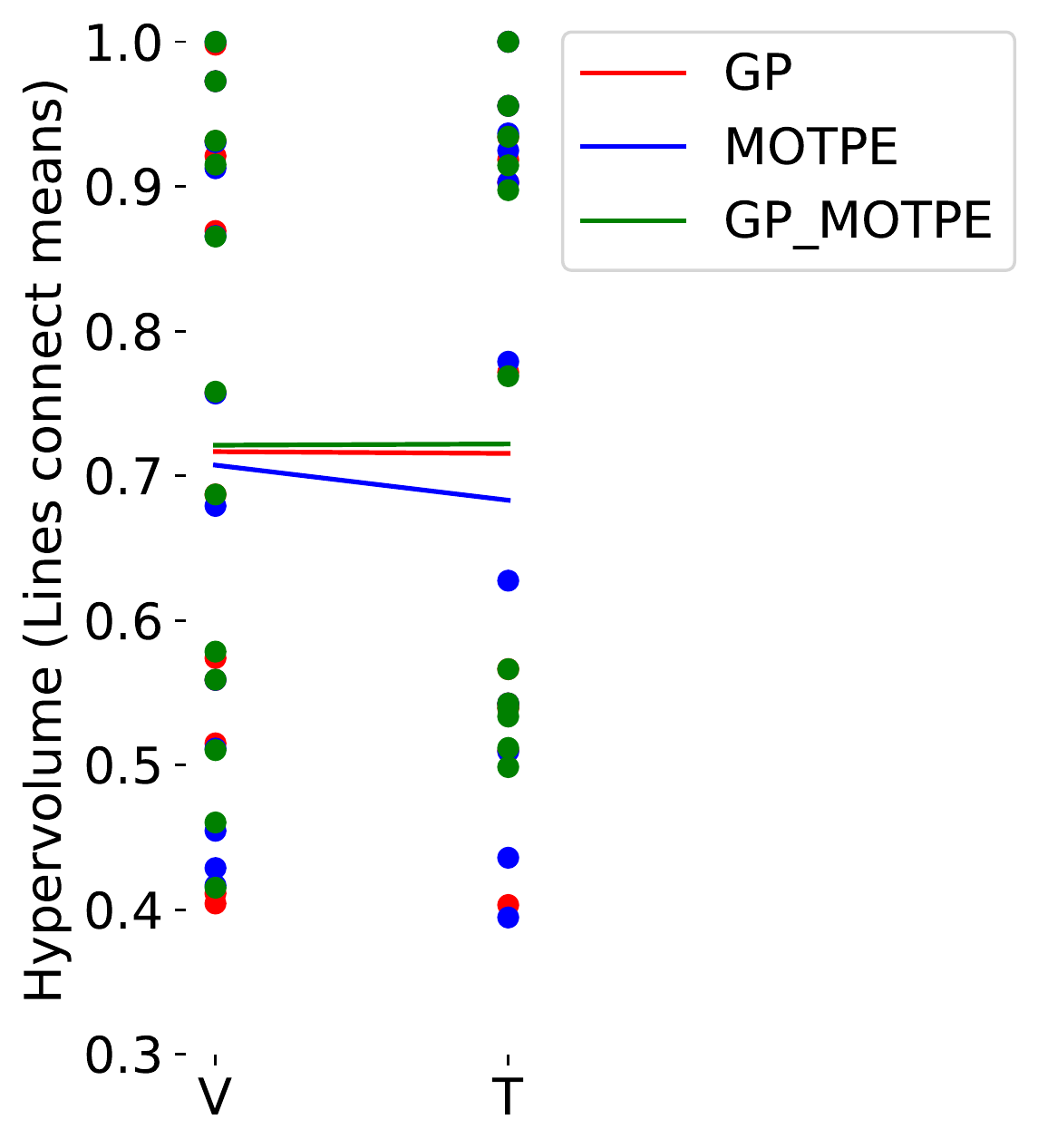}
\caption{SVM}
\label{fig:hv_vt_svm}
\end{subfigure}
\hfill
\begin{subfigure}[b]{0.3\textwidth}
\centering
\includegraphics[width=4cm]{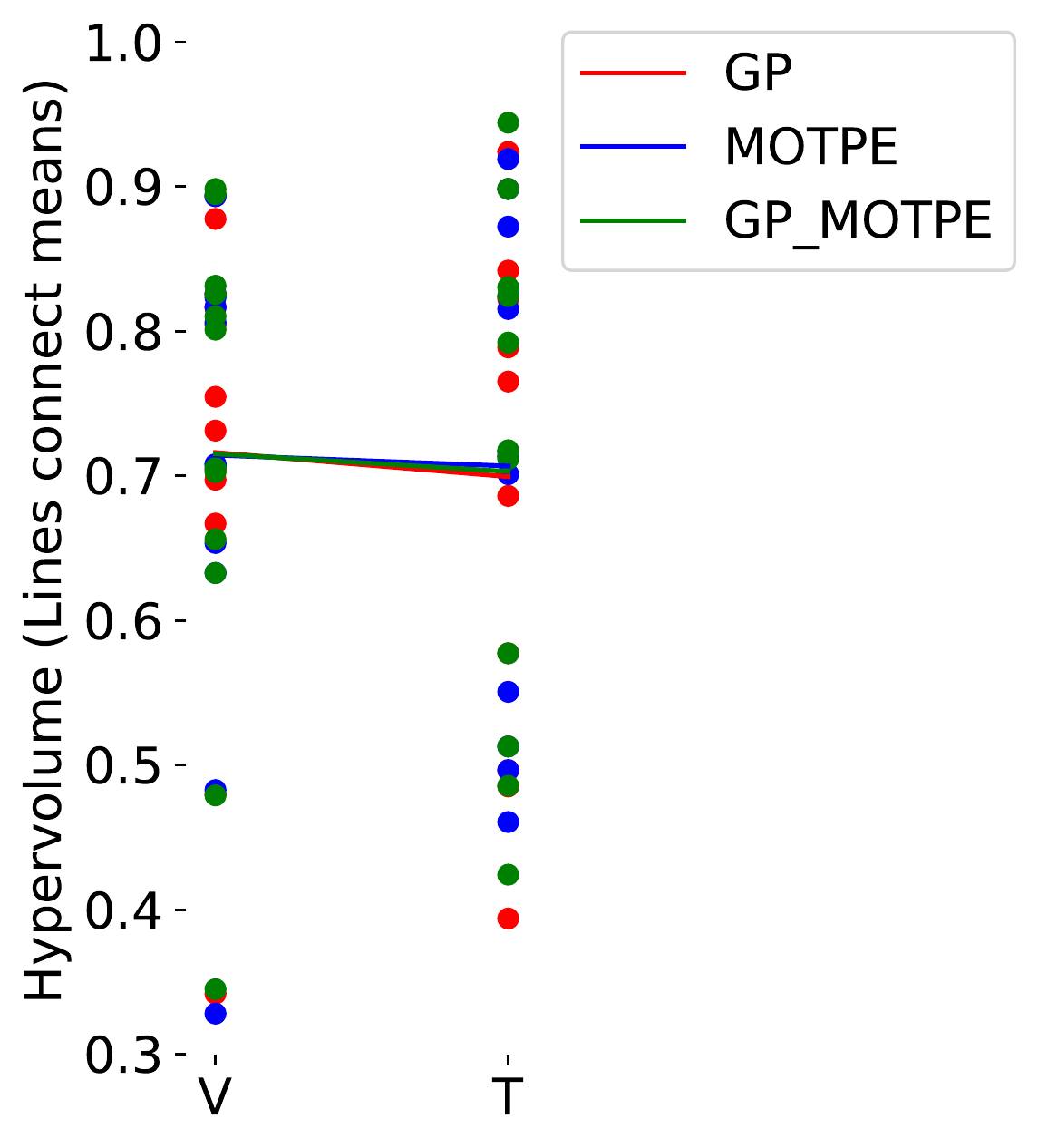}
\caption{DT}
\label{fig:hv_vt_dt}
\end{subfigure}

\caption{Hypervolume generated by the HP configurations found using the validations set (V) and then evaluated with the test set (T) }
\label{fig:hv_vt}
\end{figure*}

Once the Pareto-optimal set of HP configurations has been obtained on the validation set, the ML algorithm (trained with those configurations) is evaluated on the test set. The difference between the hypervolume values obtained from the validation and test set can be used as a measure of reliability: in general, one would prefer HP configurations that generate a similar hypervolume in the test set. Figure \ref{fig:hv_vt} shows that the difference between both hypervolume values is almost zero when GP\_MOTPE is used, for all ML algorithms. In general, MOTPE and GP\_MOTPE have the smallest (almost identical) mean absolute hypervolume difference ($0.0444$ and $0.0445$ respectively), compared with that of GP ($0.051$). However, GP\_MOTPE has the smallest standard deviation ($0.054$), followed by MOTPE ($0.066$) and GP ($0.067$).

\begin{longtable}[H]{c>{\centering}p{1cm}>{\centering}p{1.5cm}>{\centering}p{1.7cm}>{\centering}p{1.5cm}>{\centering}p{1.5cm}>{\centering\arraybackslash}p{1.7cm}}

\caption{Average rank (given by the hypervolume indicator) of each algorithm}\label{tab:HV_ML} \\

\hline  
\multirow{2}{*}{\thead{}} & \multicolumn{3}{c}{\textbf{Validation set}} & \multicolumn{3}{c}{\textbf{Test set}} \\ \cmidrule(lr){2-4}\cmidrule(lr){5-7}
 & \multicolumn{1}{c}{\textbf{GP}} & \multicolumn{1}{c}{\small \textbf{MOTPE}} & \multicolumn{1}{c}{\small \textbf{GP\_MOTPE}} & \multicolumn{1}{c}{\textbf{GP}} & \multicolumn{1}{c}{\small \textbf{MOTPE}} & \multicolumn{1}{c}{\small \textbf{GP\_MOTPE}} \\  \hline
\endfirsthead
\hline  
\multirow{2}{*}{\thead{}} & \multicolumn{3}{c}{\textbf{Validation set}} & \multicolumn{3}{c}{\textbf{Test set}} \\ \cmidrule(lr){2-4}\cmidrule(lr){5-7}
 & \multicolumn{1}{c}{\textbf{GP}} & \multicolumn{1}{c}{\textbf{MOTPE}} & \multicolumn{1}{c}{\textbf{GP\_MOTPE}} & \multicolumn{1}{c}{\textbf{GP}} & \multicolumn{1}{c}{\textbf{MOTPE}} & \multicolumn{1}{c}{\textbf{GP\_MOTPE}} \\  \hline
\endhead
\hline \multicolumn{4}{r}{{Continued on next page}} \\ 
\endfoot
\endlastfoot

\textbf{Avg. rank} & 2.125 & 1.9861 & 1.8889 & 2.1528 & 1.875 & 1.9722 \\ \hline
\end{longtable}

% \vspace{-5mm}

It is somehow surprising that the combined GP\_MOTPE algorithm does not always obtain an improvement over the individual MOTPE and GP algorithms. By combining both approaches, we ensure that we select configurations that (1) have high probability to be nondominated (according to the candidate selection strategy), and (2) has the highest MEI value for the scalarized objective. In the individual GP algorithm, (1) is neglected, which increases the probability of sampling a non-Pareto optimal point, especially at the start of the algorithm. In the original MOTPE algorithm, (2) is neglected, which may cause the algorithm to focus too much on exploitation, which increases the probability of ending up in a local optimum. We suspect that the MOTPE approach for selecting candidate points may actually be too restrictive: it will favor candidate points close to already sampled locations, inherently limiting the exploration opportunities the algorithm still has when optimizing MEI.

\section{Concluding remarks}
\label{sec:conclusions}

In this paper, we proposed a new algorithm (GP\_MOTPE) for multi-objective HPO of ML algorithms. This algorithm combines the predictor information (both predictor and predictor variance) obtained from a GPR model with heterogenous noise, and the sampling strategy performed by Multi-objective Tree-structured Parzen Estimators (MOTPE). In this way, the algorithm should select new points that are likely to be non-dominated, and that are expected to cause the maximum improvement in the scalarized objective function. 

The experiments conducted report that our approach performed relatively well for the analytical test functions of study. It appears to outperform the pure GP algorithm in all analytical instances; yet, it does not always outperform the original MOTPE algorithm. Further research will focus on why this is the case, which may yield further improvements in the algorithm. In the HPO experiments, GP\_MOTPE shows the best average rank w.r.t. the hypervolume computed on the validation set. In addition, it showed promising reliability properties (small changes in hypervolume when the ML algorithm is evaluated on the test set). Based upon these first results, we believe that the combination of GP and TPE is promising enough to warrant further research. The observation that it outperforms the pure GP algorithm (which used PSO to maximize the infill criterion) is useful in its own right, as the optimization of infill criteria is known to be challenging. Using MOTPE, a candidate set can be generated that can be evaluated efficiently, and which (from these first results) appears to yield superior results. 

\section*{Acknowledgements}
 This research was supported by the Flanders Artificial Intelligence Research Program (FLAIR).
 
\section*{Appendix 1. Setup of hyperparameters in the HPO experiments}
\label{sec:hyperparameters}

\vspace{-10mm}
\begin{longtable}[H]{p{2cm}p{6cm}>{\centering}p{1cm}>{\centering\arraybackslash}p{3cm}} \\
\hline  
\multicolumn{1}{c}{\textbf{HP}} & \multicolumn{1}{c}{\textbf{Description}} & \multicolumn{1}{c}{\textbf{Type}} & \multicolumn{1}{c}{\textbf{Range}} \\ \hline
\endfirsthead
\hline  
\multicolumn{1}{c}{\textbf{HP}} & \multicolumn{1}{c}{\textbf{Description}} & \multicolumn{1}{c}{\textbf{Type}} & \multicolumn{1}{c}{\textbf{Range}} \\ \hline
\endhead
\hline \multicolumn{4}{r}{{Continued on next page}} \\ 
\endfoot
\endlastfoot

\multicolumn{4}{l}{ \quad \quad \textit{Multilayer Perceptron (MLP)}} \\ \hline \hline
max\_iter & Iterations to optimize weights & Int. & $[1, 1000]$ \\
neurons & Number of neurons in the hidden layer & Int. & $[5, 1000]$ \\
lr\_init & Initial learning rate & Int. & $[1, 6]$ \\
b1 & First exponential decay rate & Real & $[10^{-7}, 1]$ \\
b2 & Second exponential decay rate & Real & $[10^{-7}, 1]$ \\ \hline
\multicolumn{4}{l}{ \quad \quad \textit{Support Vector Machine (SVM)}} \\ \hline \hline
C & Regularization parameter & Real & $[0.1, 2]$ \\
kernel & Kernel type to be used in the algorithm & Cat. &  [linear, poly, rbf, sigmoid]\\ \hline 
\multicolumn{4}{l}{ \quad \quad \textit{Decision Tree (DT)}} \\ \hline \hline
max\_depth & Maximum depth of the tree. If 0, then \textit{None} is used & Int. & $[0, 20]$ \\
mss & Minimum number of samples required to split an internal node & Real &  $[0, 0.99]$\\ 
msl & Minimum number of samples required to be at a leaf node & Int. &  $[1, 10]$\\
max\_f & Features in the best split & Cat. &  [auto, sqrt, log2]\\ 
criterion & Measure the quality of a split & Cat. &  [gini, entropy]\\ \hline 
\end{longtable}
\bibliographystyle{splncs04}
\bibliography{references}

\end{document}